\documentclass[twocolumn]{article}

\usepackage{amsmath}
\usepackage{amssymb}
\DeclareMathOperator{\argmax}{arg\,max}
\usepackage{todonotes}
\usepackage{algorithm2e}
\usepackage{algorithmicx}
\usepackage{algpseudocode}
\usepackage{booktabs}
\usepackage{natbib}
\usepackage{comment}
\usepackage[citecolor=blue]{hyperref}

\usepackage{authblk}

\title{Deep Relevance Regularization: Interpretable and Robust Tumor Typing of Imaging Mass Spectrometry Data}
\date{}
\author[*1]{Christian Etmann}
\author[1]{Maximilian Schmidt}
\author[1]{Jens Behrmann}
\author[1,2]{Tobias Boskamp}
\author[1]{Lena Hauberg-Lotte}
\author[1]{Annette Peter}
\author[3]{Rita Casadonte}
\author[3,4]{J\"{o}rg Kriegsmann}
\author[1]{Peter Maass}
\affil[1]{University of Bremen}
\affil[2]{SCiLS, Bruker Daltonik GmbH}
\affil[3]{Proteopath GmbH}
\affil[4]{Center for Histology Cytology and Molecular Diagnostic, Trier}
\affil[*]{To whom correspondence should be addressed}

\begin{document}

\maketitle

\abstract{Neural networks have recently been established as a viable classification method for imaging mass spectrometry data for tumor typing. For multi-laboratory scenarios however, certain confounding factors may strongly impede their performance. In this work, we introduce \emph{Deep Relevance Regularization}, a method of restricting what the neural network can focus on during classification, in order to improve the classification performance. We demonstrate how Deep Relevance Regularization robustifies neural networks against confounding factors on a challenging inter-lab dataset consisting of breast and ovarian carcinoma. We further show that this makes the \emph{relevance map} -- a way of visualizing the discriminative parts of the mass spectrum -- sparser, thereby making the classifier easier to interpret}.

\def\L{\mathcal{L}}
\section{Introduction}
\subsection{Tumor Typing with IMS data}
In recent years, imaging mass spectrometry (IMS) has seen an increased interest for applications in pathology \citep{aichler2015maldi,kriegsmann2015maldi,alberts2018maldi}. As its data is spatially resolved, connections between the morphological structures of the examined tissue and its biochemical properties can be uncovered \citep{stoeckli2001imaging}. While immunohistochemical (IHC) stainings can reliably detect certain a priori known biomarkers, the incorporation of further biochemical information may be beneficial for correctly determining the tumor type. One interesting avenue is the analysis of formalin-fixed, paraffin-embedded (FFPE) tissue, which is the standard tissue preparation and storage method in clinical pathology. Combined with IMS modalities such as MALDI-IMS (matrix-assisted laser desorption/ionization IMS \citep{maldi}), these allow for high-throughput analysis. With the rise of machine learning, processing the obtained mass spectra in an automated fashion has become feasible. In particular, the classification of these mass spectra allows for tumor \mbox{(sub-)typing} if suitable training data is available. Such training data can be annotated by a pathologist according to the tissue's morphological features, IHC stainings, the patient history and other available information, which may not even be available at test time.\\

\noindent One complication lies in technical variability of the mass spectra. These may stem from just small differences in the measurements. As each measurement involves several experimental steps (including tissue preparation), each aspect can contribute to unexpected effects within the data and add up \citep{cordero2019targeted,buck2018round}. These differences can act as confounding factors during training and thus cause a classification pipeline to classify based not only on biologically plausible features, but also on data artifacts. Any classification pipeline thus needs to exhibit robustness to these kinds of confounding factors.

\noindent This is especially relevant in many real-world scenarios, where the training data was obtained under different conditions than the data the model is applied to. In particular, this includes clinical applications for tumor typing, where the training data would typically not be created at the same time or even the same place as the data derived from the patient's tumor tissue.

\subsection{Classification of IMS data for Tumor Typing}\label{sec:classical_approach}
Classically, when constructing a classification pipeline for IMS data, one has to decide on a combination of a multitude of methods, each of which may end up having a large impact on the result:\newline
First, the data needs to be preprocessed. This may include different types of normalization (e.g. TIC normalization) \citep{deininger2011normalization}, smoothing or denoising of the spectra. Then, methods for feature extraction or feature selection need to be applied, which ensures that only the most descriptive or discriminative aspects of the data are passed on to the classifier. Often, this is accompanied by a dimensionality reduction. This is not only beneficial for numerical reasons, but may also alleviate the \emph{curse of dimensionality}, a phenomenon that impairs the performance of classifiers on data of high dimensionality (cf. \citep{friedman2001elements}). Methods for feature extraction and/or selection include supervised peak picking, non-negative matrix factorization \citep{lee1999learning,boskamp2017new,leuschner2018supervised}, principal component analysis or autoencoders \citep{thomas2016dimensionality}. These features are then passed to an appropriate classifier, such as linear discriminant analysis classifiers (LDA), support vector machines (SVM), neural networks (NN) or random forests \citep{boskamp2017new,galli2016machine}.\newline
Since the number of possible combinations of these steps is very large, constructing a good classification pipeline is a very challenging task. This is impeded by the fact that most methods require some form of hyperparameter tuning.\newline
Another important aspect lies in the interpretability of the obtained classification pipeline. In order to be accepted by doctors and patients alike, a biologically plausible and interpretable model is desirable, which allows for further validation besides classification accuracies. If the mass spectra are e.g. classified with a simple linear classifier based on a subselection of peaks (e.g. based on AUC-values), the classifier's weights provide information which peaks the classifier takes into account the most. This can then be checked against a priori known cancer biomarkers. If on the other hand the feature extraction is performed with a kernel PCA \citep{scholkopf1998nonlinear} and the resulting features are then classified using a perceptron \citep{rosenblatt1958perceptron}, a verification of the biological plausibility of this pipeline is much more difficult, since neither the features nor the classifier allow for a direct interpretation.

\subsection{Tumor Typing using Deep Learning}\label{sec:DL_approach}
\begin{figure}
    \centering
    \includegraphics[width=.45\textwidth]{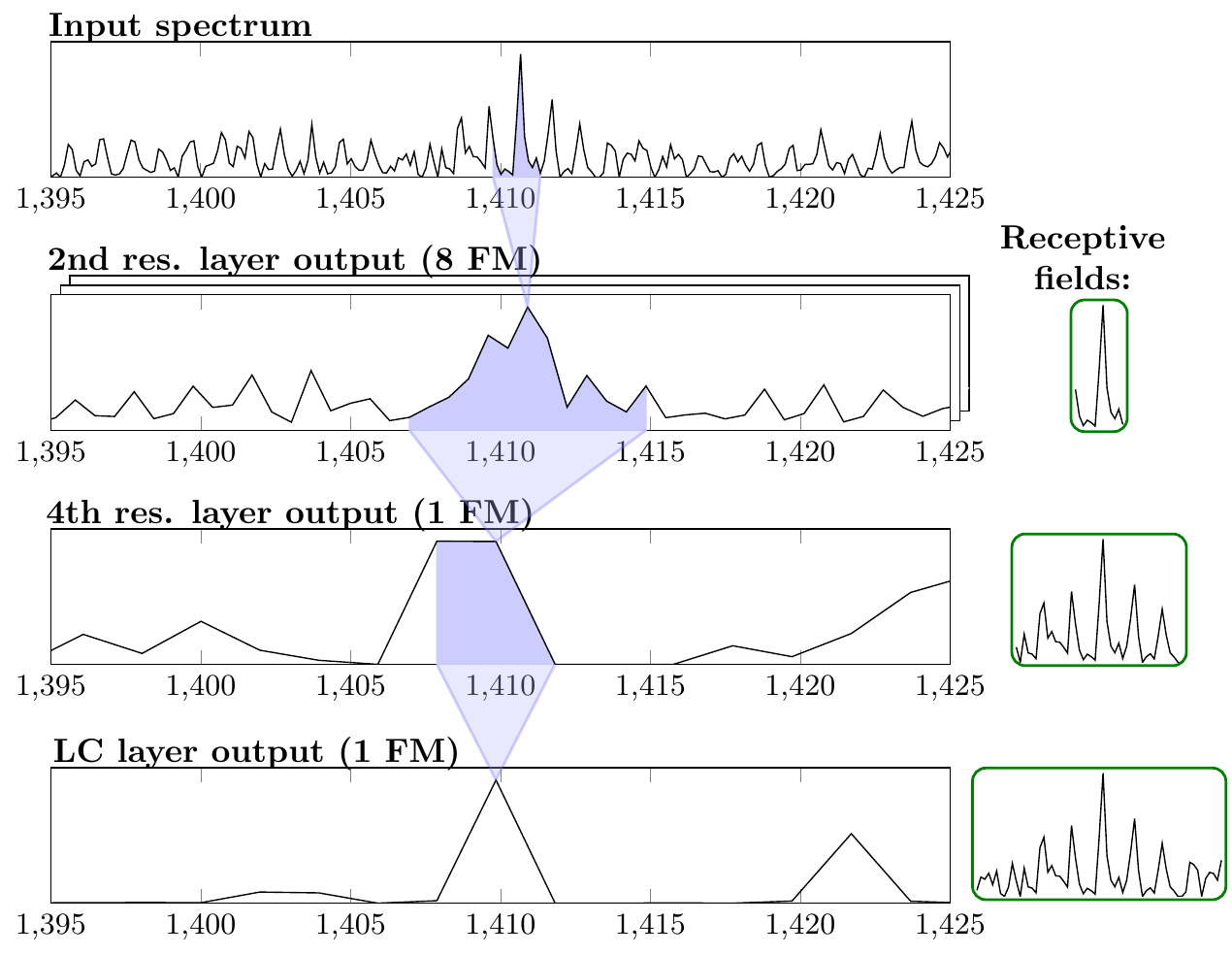}
    \caption{IsotopeNet applied to a mass spectrum of a lung tumor, from \citep{isotopenet}. The convolutional and locally-connected layers are only applied as many times as necessary for the receptive fields to cover the isotopic patterns of peptides, based on the 'averagine' model \citep{senko1995determination}. Afterwards, a fully-connected layer translates these to class membership scores.}
    \label{fig:isotopenet}
\end{figure}

Another concept for tumor typing based on IMS data can be realized through \emph{Deep Learning} \citep{lecun2015deep}, where a deep NN is trained in an end-to-end fashion directly on raw (or normalized) mass spectra \emph{without} performing any separate feature extraction. Neural networks consist of a cascade of parametric transformations (called the \emph{layers} of the NN), which are then jointly optimized with a final logistic regression layer for classification. In a sense, this means that the neural network may be able to automatically learn the optimal feature extraction for the classification task at hand (with respect to the chosen network architecture). In \citep{isotopenet}, a suitable neural network architecture named \emph{IsotopeNet} was introduced, as visualized in Figure \ref{fig:isotopenet}. Through convolutional layers \citep{lecun1989backpropagation}, it extracts local structures (such as the (relative) heights and positions of neighboring ion abundances, peaks and isotopic patterns), which are then related to their position on the mass spectrum.\newline
%Neural networks are often seen as opaque 'black box models', whose classification decision cannot be retraced. In contrast to this, various approaches have surfaced in the last years, which allow for an assessment of the most salient regions in the input for the respective classification. When applied to a mass spectrum, they return a list of scores for estimated evidence or counter-evidence with respect to a class for each m/z-value.

\section{Materials and Methods}
\subsection{Neural Networks for IMS Data}
Here, we very briefly review neural networks and their application to classification for IMS data. For more in-depth information, interested readers are referred to \citep{isotopenet,lecun2015deep,Goodfellow-et-al-2016}.\newline
For a problem with $C$ classes (which will be set to $C=2$ in the later experiments), a neural network for classification can be described as a function $f_\theta : \mathbb{R}^n \rightarrow \mathbb{R}^C$, which assigns a score $f_\theta(x)$ to a mass spectrum $x \in \mathbb{R}^n$. For MALDI-TOF data, typically $n=10^5 \dots 10^6$. The classification is performed according to the highest score, i.e.
$$F_\theta(x) = \argmax_i f_\theta(x)_i.$$ The function $f_\theta$ is often a composition of parametric functions with parameter $\theta$, i.e. $f_\theta = g_L \circ \dots \circ g_1$, but other network topologies are also common. Each of these functions is called a \emph{layer}. A prototypical example of such a layer is a \emph{convolutional layer} defined by 
\begin{equation}
g: x \mapsto \text{ReLU}(K \ast x + b),
\end{equation}
where $K \ast x$ represents the convolution of $x$ with a \emph{filter kernel} $K$, $b$ is a \emph{bias vector} and $\text{ReLU}$ is a certain nonlinear function, the \emph{rectified linear unit} \citep{He_2015_ICCV}. $K$ and $b$ are the trainable parameters of this layer. The trainable parameters of the whole network are then represented by the vector $\theta$.\newline
In the case of classification networks, the output of the network is converted to a distribution of class membership probabilities via the softmax-function given by
$$\text{softmax}: z \mapsto (p_1,\dots,p_C)^T, \text{ where } p_i=\tfrac{e^{z_i}}{\sum_{k=1}^C e^{z_k}}.$$ We write $p_\theta(x):=\text{softmax} (f_\theta(x))$ and call $z=f_\theta(x)$ the logit-vector of the spectrum $x$. The network is trained by employing the principle of empirical risk minimization (ERM). This amounts to minimizing 
\begin{equation}\label{eq:erm_vanilla}
    \frac{1}{N} \sum\limits_{i=1}^N \mathcal{L}\left(p_\theta(x^{(i)}),y^{(i)} \right),
\end{equation}
w.r.t. $\theta$, where $\mathcal{L}$ denotes the negative log-likelihood loss function and where $\{(x^{(i)},y^{(i)})\}_{i=1,\dots,N}$ is the training set of labelled mass spectra. This loss is typically minimized with stochastic gradient descent or modifications thereof, such as Adam \citep{adam}. \newline
The IsotopeNet-architecture, while optimized for the specific structure of IMS data, follows the principles presented here.

\subsection{Interpretation of Classification Results}\label{sec:attribution_methods}
In order to be accepted as part of the medical diagnosis toolbox, an automated classification pipeline needs to allow for some degree of interpretability. While some models, e.g. linear models, exhibit \emph{inherent interpretability} via inspection of their weights, black-box models such as neural networks do not allow for this. Nevertheless, a multitude of \emph{attribution methods} for neural networks have been developed, which allow for a \emph{post hoc} interpretation by providing some estimate of feature importance for the classification. The \emph{relevance map} $\rho_\theta(x,y) \in \mathbb{R}^n$ should exhibit large positive values at the parts of the input $x$ that act as strong evidence for class $y$, whereas large negative values should signify strong counterevidence against class $y$. Values close to $0$ should play little to no role in the classification. \newline
Attribution methods roughly fall into one of two categories: perturbation-based attribution methods and backpropagation-based attribution methods. Perturbation-based methods such as Occlusion \citep{zeiler2014visualizing} and Prediction Difference Analysis \citep{zintgraf2017visualizing} observe the influence on the classification score after perturbing different parts of the input. Backpropagation-based methods typically calculate their relevance attribution via modifications of gradient backpropagation in the neural network. Among these are \emph{saliency maps} \citep{simonyan2013deep} (unmodified gradients), the misnomered \emph{deconvolution} \citep{zeiler2014visualizing}, {guided backpropagation} \citep{springenberg2014striving}, {layer-wise relevance propagation} \citep{bach2015pixel} and its generalization {deep Taylor decomposition} \citep{montavon2017explaining, montavon2017methods}, {DeepLIFT} \citep{shrikumar2016not}, {forward-backward interpretability} \citep{balu2017forward}, {VisualBackProp} \citep{bojarski2016visualbackprop}, Excitation Backprop \citep{zhang2016top}, GradCAM \citep{selvaraju2016grad} and PatternNet/PatternAttribution \citep{kindermans2017patternnet}. %There are also 'meta' attribution methods, which calculate averages or line integrals of the above-mentioned attribution methods over regions of the input space, such as SmoothGrad \citep{smilkov2017smoothgrad}, and integrated gradients \citep{sundararajan2017axiomatic}. For image data, these are known to result in better-looking relevance maps, but require many evaluations.\\

\noindent In \citep{isotopenet}, a first step towards interpretable neural networks for IMS data was taken. The authors propose to inspect $\nabla_x f_\theta^y(x)$, the gradient of the logit prediction score with respect to the class $y$. This is the aforementioned \emph{saliency map}, as it is often called in the field of computer vision. The gradient represents the input-output-sensitivity and thus serves as a simple attribution method, which parts of the input spectrum influence the classification the most. The authors further identify a problem with this approach: Since $$\nabla_x f_\theta^y(x) = \left( \frac{\partial f_\theta^y(x)}{\partial x_1},\dots, \frac{\partial f_\theta^y(x)}{\partial x_n} \right)^T,$$ each entry represents the change in $f_\theta^y(f)$ per unit of $x_i$, i.e. per a certain number of measured ions (up to linearization). A fixed difference in the number of measured ions may mean a very large \emph{relative} change for a rare ion, but a miniscule change for a very abundant ion. The authors thus propose to multiply each entry by its standard deviation on the training set as a crude measure of the 'typical' rate of change of ions at this m/z-position.

\noindent Here, we choose a different approach that is theoretically well-founded and which ends up being conceptionally very similar, the above-mentioned \emph{layerwise relevance propagation} (LRP). While LRP has a few varieties, we choose a variant (\emph{z-rule}) that allows for easy implementation in order to encourage adoption by researchers and practitioners. Propagating from the value of an output neuron back to the input, LRP uses a conservation law in order to assign a neuron's relevance according to its additive contribution to its output. As \citep{ancona2017towards}, \citep{kindermans2016investigating}, \citep{shrikumar2016not} show, for neural networks with exlusively ReLU nonlinearities, LRP reduces to 
\begin{equation}\label{eq:epsilon-LRP}
\rho_\theta(x,y) = x\odot \nabla_x f^y_\theta(x)\\
\end{equation}
(where $\odot$ denotes the entry-wise multiplication), which is the case e.g. for IsotopeNet and many other neural network architectures.

\subsection{Deep Relevance Regularization}
For IMS data used in tumor classification tasks, it can often be assumed that only a small fraction of each input spectrum is actually relevant in the classification. Due to artifacts induced during the tissue preparation or the spectral measurement (such as delocalization or ion suppression effects \citep{cole2011electrospray}), some region of the mass spectra may, however, correlate with a certain class, despite not actually being a biomarker. This may e.g. occur, if the test data was recorded with a different machine or by a different operator than training data. A low number of training samples aggravates this problem, since this makes distinguishing noise from signal more difficult. \newline
Ideally, we would like to restrict our model to only take into account the most \emph{relevant} m/z-values, while ignoring the above mentioned data artifacts. One such restricted model can be regarded as simpler and may thus be expected to be more robust to overfitting.\newline
In other words, for the classifier to only take into account the most important parts of $x$, the vector $\rho_\theta(x,y)$ should generally be \emph{sparse}, i.e. have most entries close to zero. To this end, we modify our training objective to be
\begin{equation}\label{eq:minimizer_general}
\min_\theta \frac{1}{N} \sum\limits_{i=1}^N \left[ \L (p_\theta(x^{(i)}),y^{(i)}) + \mathcal{R}(\lambda, \rho_\theta(x^{(i)},y^{(i)})) \right],
\end{equation}
where the \emph{regularization term} $\mathcal{R}(\lambda, \rho_\theta(x,y))$ assumes low values for sparse relevance maps and high values for non-sparse (\emph{dense}) relevance maps. Here, $\lambda$ controls the regularization strength. The formulation in \eqref{eq:minimizer_general} ensures that the resulting classifier $f_\theta$ strikes a balance between accuracy and relevance sparsity. We call this proposed approach \emph{Deep Relevance Regularization} (DRR). For sufficiently overparameterized model families (like most NNs), DRR can be regarded as a model selection mechanism, in which case one might hope to find a model which does not even reduce the accuracy at all compared to the unregularized network. This approach is conceptionally similar to \citep{ijcai2017-371} and \citep{rieger2019interpretations}. Generally, a sparser relevance map can be expected to be easier to interpret, because fewer m/z-values would need to be tested for their biological relevance.\\

\noindent A natural measure for sparsity is the '0-norm' $\|\rho_\theta(x,y)\|_0$, which counts the non-zero entries of $\rho_\theta(x,y)$. The 0-norm is unsuited for optimization purposes, however, because it has derivative 0 almost everywhere and exhibits jump discontinuities. Furthermore, it may be too restrictive for our purposes, since a very small entry of $\rho_\theta(x,y)$ is given the same weight as a very large entry. We therefore use the 1-norm as a proxy measure for sparsity. The sparsity-promoting property of the 1-norm is well known in areas such as inverse problems, compressed sensing and machine learning. One example is the LASSO for linear regression \citep{tibshirani1996regression}, where the loss function
\begin{equation}
	\|Y-WX\|_F + \lambda \|W\|_1
\end{equation}
is minimized\footnote{Here, $\|\cdot\|_1$ denotes the entrywise 1-norm, not the operator 1-norm.} with respect to $W$ with data matrices $X$ and $Y$, which leads to sparse $W$. However, a well-known phenomenon that arises in LASSO models is that out of a group of highly correlated patterns, only one (or few) are selected by $W$. This can be explained by the model gaining additional sparsity through dropping this supposedly 'redundant' information. If this is not desired, the effect can be mitigated by adding an additional regularization term $\mu \|W\|^2_F$ to \eqref{eq:minimizer_general}, where $\mu \geq 0$. The resulting model is then called the \emph{elastic net} \citep{zou2005regularization}. In trypsin-digested IMS data, highly correlated patterns are found on several levels: Each peak is several m/z-bins wide, isotopic patterns consist of several neighboring peaks and peptide patterns contain correlated patterns of different peptides. If one wishes to include this information for DRR, a similar regularization term should be employed. This also offers advantages for interpretability, since this information is retained. For these reasons, in the following we choose 
\begin{equation}\label{eq:minimizer_elastic}
\mathcal{R}(\lambda,\rho_\theta(x,y))=\lambda_1 \|\rho_\theta(x,y)\|_1 + \lambda_2 \|\rho_\theta(x,y)\|^2_2
\end{equation}
to be the DRR term in the objective \eqref{eq:minimizer_general} for $\lambda_1,\lambda_2\geq 0$.\\

\noindent Whether the 'surviving' m/z-values are actually biologically relevant or themselves confounding factors, is to be determined empirically. We however expect that e.g. differing baselines between the classes should result in large intervals of the spectrum being deemed 'relevant' for the classification, in contrast to isolated peaks, which can be expected to be the result of specific measured ions.

\subsubsection{Logit or Softmax Scores}
It is generally advisable to use logit scores instead of softmax scores for the calculation of the LRP-term \eqref{eq:epsilon-LRP}. This is because for softmax scores, the gradients (and thus the relevance maps) tend to 0 as class prediction probabilities tend to either 0 or 1. This is especially striking in the 2-class case. Let $(p_1,p_2)^T:= p_\theta(x)$ and $(z_1,z_2)^T:=f_\theta(x)$, then 
\begin{align*}
\nabla_x p_1 &= p_1(1-p_1) \left( \nabla_x z_1  - \nabla_x z_2 \right)\\
\nabla_x p_2 &= \nabla_x (1-p_1)= -\nabla_x p_1.
\end{align*}
When using this as the relevance function for DRR, this has the adverse effect that the samples with probabilities close to 50\% are penalized more strongly than those that the NN is 'surer' about. However, for the already well-classified samples \emph{in the training set}, one can 'afford' a more restrictive model (i.e. with a higher imposed sparsity on the relevance), whereas the badly-classified examples require more flexibility.\newline
Note that this logic is reversed for \emph{test data}: For badly-classified examples in the test set, a more restrictive model (i.e. higher $\lambda_1$, $\lambda_2$) can be expected to yield a better classification for spectra that are difficult to classify. What are the 'right' values of $\lambda_1$ and $\lambda_2$ can however not be assessed purely by training on the training set, because performance on the training set tends to be highest for the most unrestricted models. One should therefore select these values based on a (cross- or hold-out) validation procedure.

\subsubsection{DRR and Weight Regularization}
One might also consider applying a sparsity penalty to the weights instead of the relevance map. For multilayer perceptrons (MLPs, i.e. NNs that consist only of fully-connected units), this is a sensible approach, as apprioriately-set zeroes in the first layer will eliminate the influence of certain entries of $x$. In most common situations, as in the case of e.g. IsotopeNet, this is however not adequate:
\begin{itemize}
	\item For high-dimensional data (like IMS data), the number of weights of deep NNs may increase drastically. Apart from memory restrictions, this also increases the likelihood of overfitting. Furthermore, these do not make use of localized information, for which convolutional layers are more suited, as explained in \citep{isotopenet}.
	\item In convolutional layers, parameter sparsity implies neither sparse gradients nor activations: For example, a very large convolutional kernel with zeroes everywhere except for the entry 1 in the middle, while being highly sparse, realizes exactly the identity function\footnote{For zero-padded convolutions.}.
	\item Residual connections \citep{he2016deep} allow for the training of very deep NNs and are ubiquitous in classification architectures (such as IsotopeNet). Here, penalizing the norm of weights lead to a mapping that is close to the identity function.
\end{itemize}
For the last two points, however, certain activation functions like ReLU may affect the sparsity as well. Another big advantage of DRR over weight sparsity is that of higher flexibility: For MLPs, a zero in the first hidden layer's weight matrix excludes this m/z-position for every examined mass spectrum. DRR on the other hand allows the model to assign a low relevance to a certain position for some mass spectrum, whereas the model may assign a high relevance to it for a different spectrum. In other words, the neural network may individually 'decide' whether a certain peak is relevant for the classification of a mass spectrum.

%\subsubsection{Complexity Analysis}
%Here, we briefly describe the time and space complexity of a DRR-regularized neural network. Assuming a vanilla neural network with $L$ layers \emph{without DRR}, the forward propagation as well as the backpropagation take $2L-1$ operations in total (assuming equal computational burden for the forward and backpropagation through each layer). 
%As DRR includes gradients in the loss \eqref{eq:minimizer_general}, the calculation of the weight-gradients for stochastic gradient descent results in so-called \emph{double backpropagation}. As describe in \textbf{(Drucker, LeCun)}, one can picture the calculation of $\nabla_x f^y_\theta(x)$ for the DRR as an appended network to the original network.\\

\subsection{Data Aquisition}\label{sec:data_description}
\subsubsection{Tissue Preparation}
Here, we give a brief description of the used dataset, which is a subset of the data used in \citep{cordero2019targeted}. Readers interested in the minutiae of the data acquisition protocols are referred to the original publication.\newline
FFPE tissue samples from breast carcinoma ($N = 99$ patients, all human epidermal growth factor (Her2) positive) and ovarian carcinoma ($N = 84$ patients, various kinds) were kindly provided by the University Hospital Heidelberg in accordance with the regulations of the local ethics committee. The cancer biopsies were assembled to four tissue microarray (TMA) blocks as cylindrical tissue cores with 1 mm diameter. Two of these TMAs consisted solely of breast tissue, while the other two consisted solely of ovarian tissue\footnote{Note that \citet{cordero2019targeted} used 5 TMA blocks. We discarded one TMA of ovarian tissue samples in order to take the stark class imbalance out of the equation.}. Slices of 5 $\mu$m thickness were washed in an ethanol series and antigen retrieval was performed in a Tris buffer. Tryptic digestion was performed on-tissue. CHCA matrix was applied with an ImagePrep device (Bruker Daltonik, Bremen, Germany).

\subsubsection{IMS Measurement}
The same TMAs were measured in two different laboratories, one in Bremen (HB) and one in Trier (TR). In both cases, the measurements were performed with an autoflex speed MALDI mass spectrometer (Bruker) in positive-ion mode. Mass spectra were collected with 100 $\mu$m spacing between spot centers. An external calibration was performed using Peptide Calibration Standard II (Bruker). Mass resolving power was approx. R = 11 000. The mass accuracy was visually estimated to be approx. 50 -- 100 ppm. 

\noindent Afterwards, the matrix was removed with 100\% methanol, which was followed by an H\&E staining. A pathologist annotated regions of high tumor concentration, which were transferred to the recorded MALDI measurement on the level of single IMS spots and henceforth used as labels for the data analysis. Note that this was done separately for each section measured in the two laboratories, such that their annotated regions differ. After this procedure, the HB measurement consisted of 5230 spectra of breast tumors and 6777 spectra of ovarian tumors. For TR, 3479 points were assigned the label 'breast' and 4621 were assigned the label 'ovary'.

\subsubsection{Preprocessing}
A baseline correction of the MALDI IMS data was performed using SCiLS Lab (version 2017a, SCiLS, Bremen, Germany) with default settings. Next, the data was imported into MATLAB (version 2018a, MathWorks, Natick, MA, USA), and for this study reduced to the m/z-range 800 Da -- 2000 Da. The spectra were subsequently normalized by total ion count.

\section{Experiments}
In the following, we will describe the experiments performed on the dataset described in section \ref{sec:data_description}. In order to simulate realistic, real-world clinical applications, we will never report test scores on the same patients or the same lab as the data the model was trained on. Along with the fact that the breast and ovarian tissue were placed on different TMAs, this carries the danger of a neural network learning biologically irrelevant factors.\newline
As a performance measure for this binary classification problem, we used the \emph{(class) balanced accuracy} $\text{balAcc} = \frac{1}{2} \left(\frac{\text{TP}}{\text{TP+FN}} + \frac{\text{TN}}{\text{TN+FP}} \right)$, where TP/(TP+FN) denotes the true positive rate and TN/(TN+FP) denotes the true negative rate (with respect to class 'breast'). This measure is not biased by the proportions of class abundance in the data, unlike the accuracy (i.e. the fraction of correctly classified samples).\\

\noindent We first train baseline neural networks without DRR, which we then compare to DRR-regularized networks. The regularization strength for DRR is chosen based on a cross-validation (CV) procedure within the same lab. For a fair comparison with other established methods, we further train a linear model on a peak-picking approach, where the number of peaks is determined according to the same hyperparameter search strategy as in the case of DRR. This was e.g. used in \citep{boskamp2017new,isotopenet,leuschner2018supervised} as well as a baseline.

\subsection{Unregularized Networks}
In order to obtain the above-mentioned baseline to compare our model against, we first train a vanilla IsotopeNet \citep{isotopenet} without DRR. To this end, we employ an \emph{inter-lab cross-validation procedure}, where we divide the patients randomly into 5 roughly equally-sized groups. For every cross-validation configuration, we then train on 4 folds of one lab and afterwards classify the \emph{other lab's} spectra of the patients \emph{not used for training}. This is done for both labs. The procedure is visualized in Figure \ref{fig:cv}(a). This 5-fold CV procedure for 2 labs results in a total of 10 neural networks trained for the unregularized baseline model. As the predicted labels form disjoint sets, we can calculate the balanced accuracy for their disjoint union.\newline
Since our target is a high class-balanced accuracy, one should take the class proportions into account in the training procedure, such that members of rare classes are assigned more weight than members of a frequent class. 
For a spectrum-label pair $(x,y)$, we weigh its loss $\mathcal{L}(p_\theta(x),y)$ in the ERM-term \eqref{eq:erm_vanilla} inversely proportional to the relative abundance of class $y$. We normalize this by the number of classes, such that the original ERM \eqref{eq:erm_vanilla} is recovered for balanced classes. This resulting class-weighted ERM-term is thus written as
\begin{equation}\label{eq:erm_class_weighted}
    \frac{1}{N} \sum\limits_{i=1}^N w_y \cdot \mathcal{L}\left(p_\theta(x^{(i)}),y^{(i)} \right),
\end{equation}
with $w_y = \tfrac{1}{C}\cdot\tfrac{N}{N_y}$, where $N_y$ denotes the number of samples of class $y$ in the respective training set. This is a common approach, for example utilized as the standard option for class weighting in \texttt{scikit-learn}, a popular machine learning library in Python \citep{scikitlearn}. The calculations were performed in our MATLAB-library \texttt{MSClassifyLib}, which serves as an API to \texttt{Tensorflow 1.12}, using an NVIDIA GeForce GTX 1080Ti GPU.\\

%\noindent The above-described procedure resulted in a (spot-wise) balanced accuracy of just 30.2\% on the data of the HB laboratory and 48.0\% on TR. When aggregating over the patients (i.e. when assigning a single label to each patient according to a majority vote over the their spots), the balanced accuracies drop even further to 26.3\% (HB) respectively 48.3\% (TR). As these results are worse than guessing on average, they suggest that the neural networks learn confounding factors instead of biologically relevant properties.\\

\noindent\textbf{Results}\newline
\noindent The above-described procedure resulted in a (spot-wise) balanced accuracy of just 37.3\%. When aggregating over the patients (i.e. when assigning a single label to each patient according to a majority vote over the their spots), the balanced accuracy drops even further to 34.9\%. As these results are worse than guessing on average, they suggest that the neural networks learn confounding factors instead of biologically relevant properties.\\

\begin{figure}
    \centering
    \includegraphics[width=.45\textwidth]{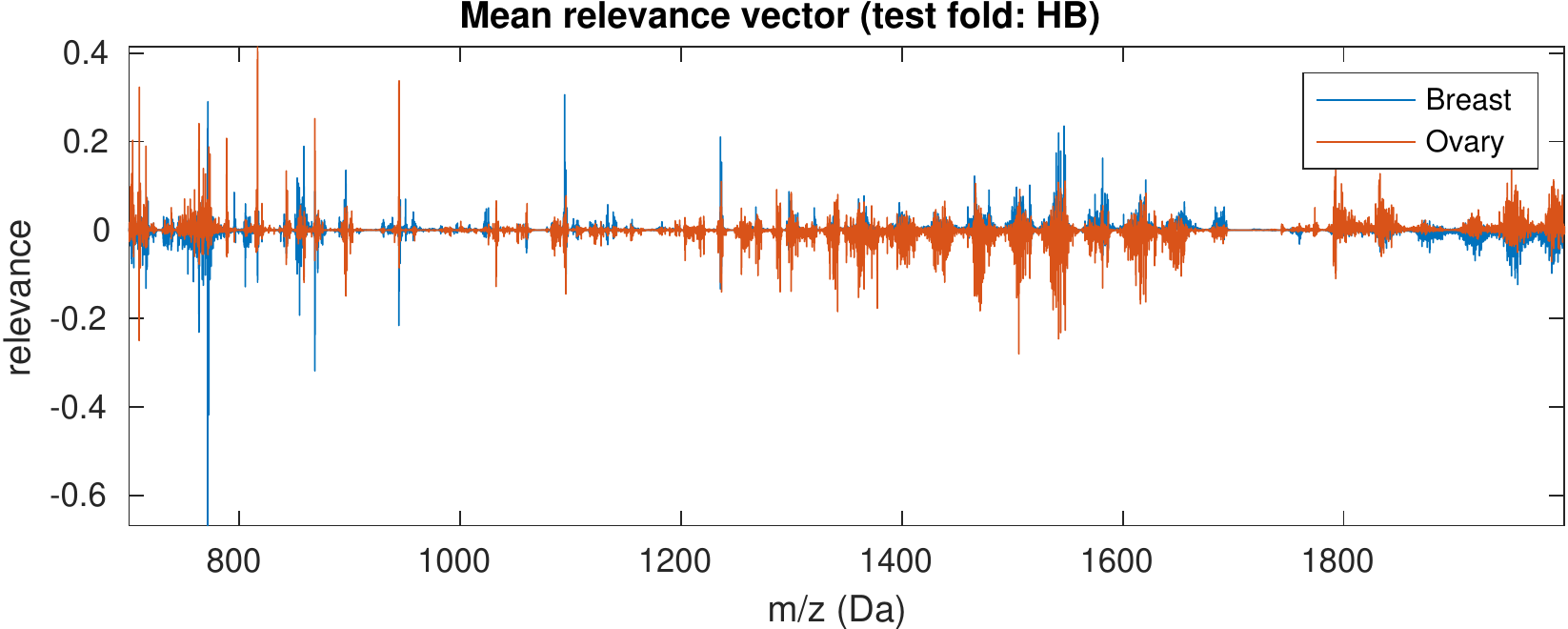}
    \caption{Mean relevance maps per class of one of the five test folds of the HB lab. Large intervals of seemingly relevant peaks instead of isolated peaks indicate that confounding factors (e.g. baselines) are being learned.}
    \label{fig:bad_relevance}
\end{figure}

\noindent In order to check this, we make use of layerwise relevance propagation. In Figure \ref{fig:bad_relevance}, the average relevance map with respect to the respective correct class of one of the test sets (on HB) is visualized. While a few singular peaks appear, the neural network mostly seems to regard larger intervals as relevant. This points towards confounding factors (e.g. differing baselines) being learned instead of actual biomarkers.\newline
We emphasize that we decided to visualize the relevances with respect to the logit of the \emph{correct class}. While the logit of the other class may be larger (resulting in a misclassification), both relevance maps may be inspected independently.\\

\noindent We further tried out weight decay as a regularization strategy with different parameter values, which was not successful either. This underlines that this is not a problem of overfitting (induced by using too little training data), but a true distribution shift between the training and test data. Furthermore, there may be spurious correlations between the two classes due to them being present on separate TMAs.

\subsection{Application of DRR}
\begin{figure}
\begin{tabular}{cc}
\includegraphics[width=.2\textwidth]{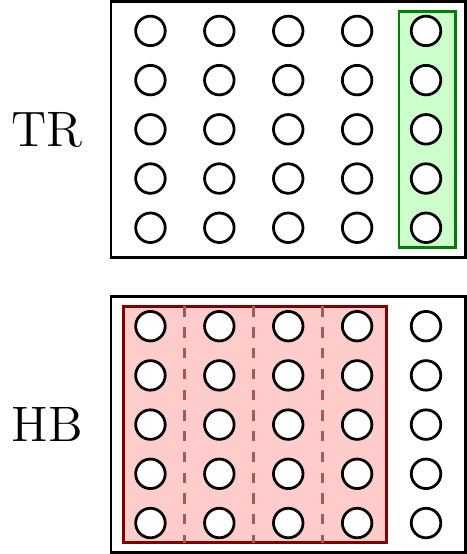}\hspace{.05\textwidth} & \includegraphics[width=.2\textwidth]{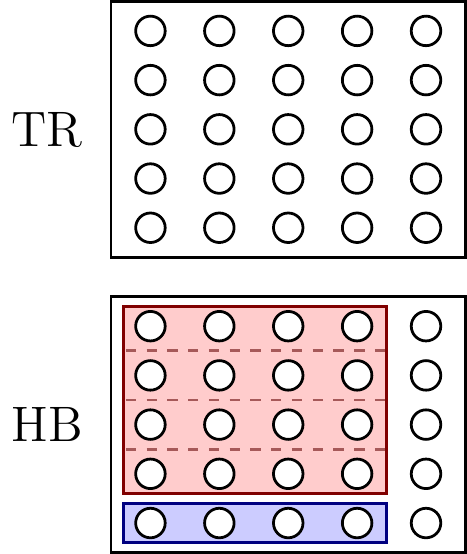}\\
\scriptsize (a) Outer CV &\scriptsize (b) Inner CV
\end{tabular}
\caption{Visualization of the inter-lab nested cross-validation procedure. For the outer CV, the models are trained on one lab (shown in red), but tested on the data from the remaining patients recorded in the other lab (shown in green). In order to choose a good hyperparamter for the DRR, an inner CV is employed, where the model is tested on the same lab (shown in blue). This is done for both labs.}\label{fig:cv}
\end{figure}

\begin{figure}
    \centering
    \includegraphics[width=.23\textwidth]{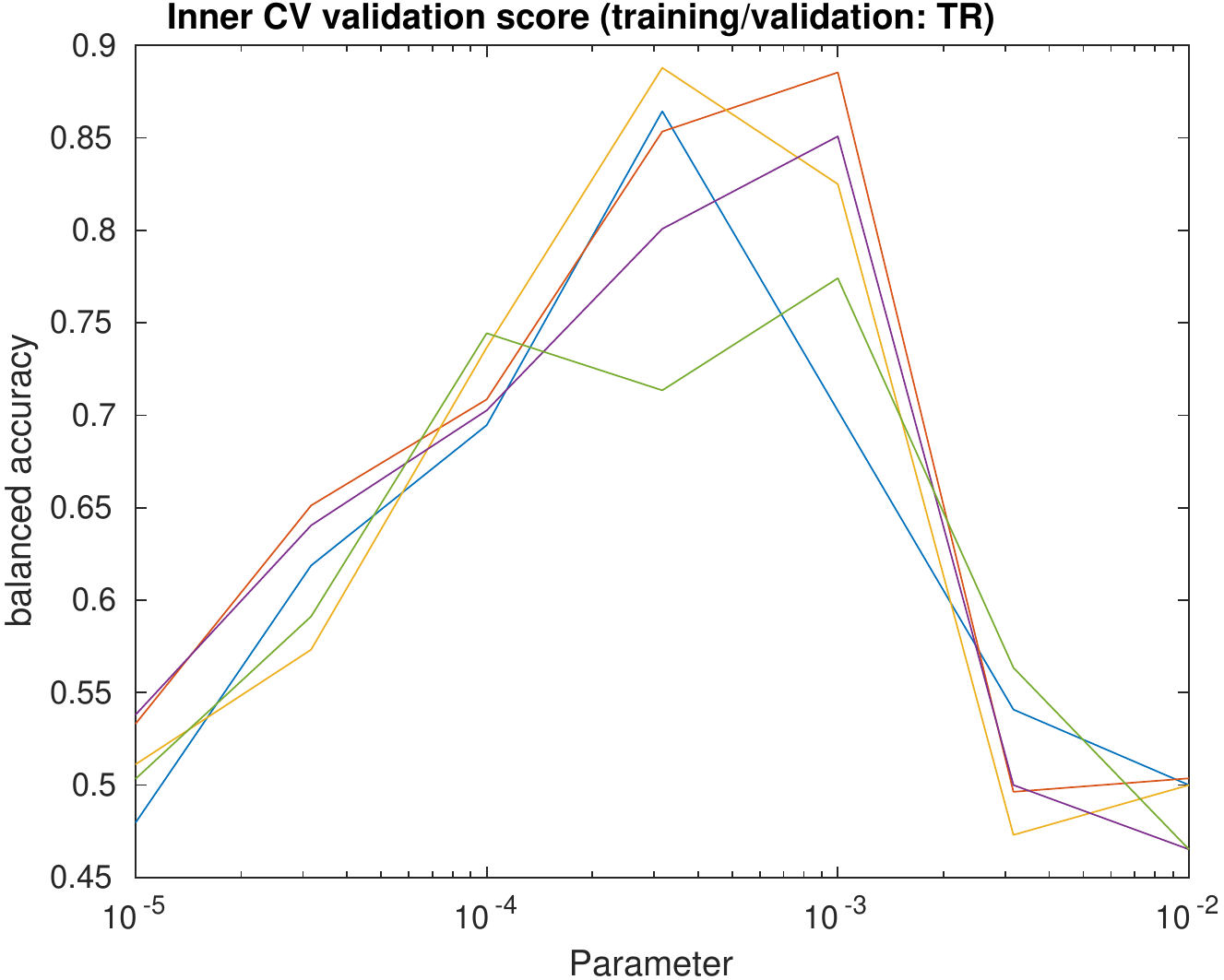}
    \includegraphics[width=.23\textwidth]{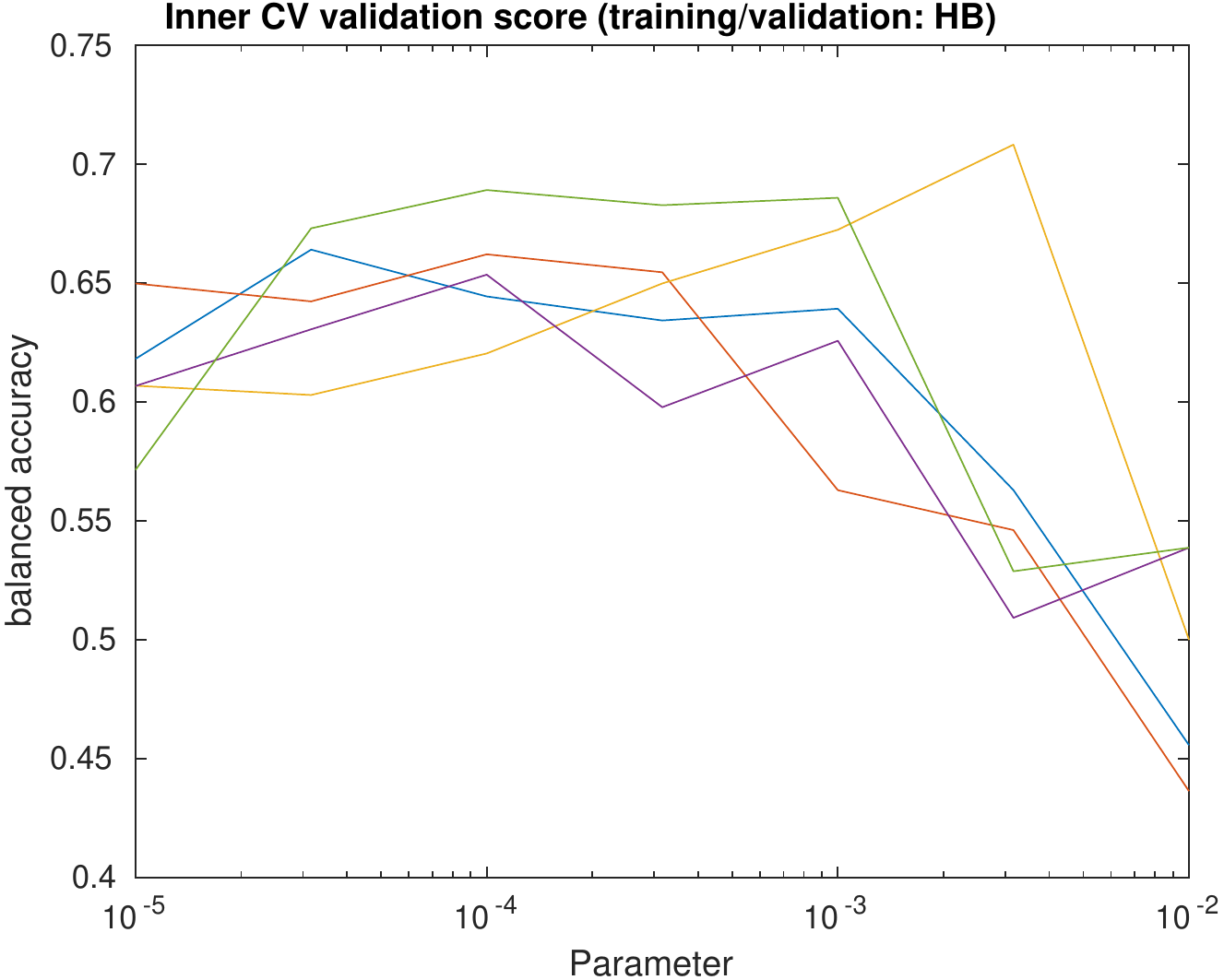}
    \caption{Parameter choice for different validation folds of the inner CV. While on HB, the performance is a little more consistent than on TR, the drop in balanced accuracy towards the upper and lower ends indicate a mostly suitable choice of parameter range. In practice, one could try to determine this range and choose the mean on the logarithmic scale as a simple heuristic.}
    \label{fig:parameter_choice_DRR}
\end{figure}

We now repeat the experiment using DRR, with otherwise equal settings. Our goal is to find out whether a neural network with DRR trained on data from one lab performs well on another lab's data. For this reason, a fair hyperparameter selection strategy should be employed, which \emph{only} uses information from the training lab. We therefore employ a \emph{nested inter-lab cross-validation strategy}, which is visualized in Figure \ref{fig:cv}. The outer CV corresponds to the same CV as for the unregularized network. Every training set of the outer CV is again randomly divided into 5 folds (based on patients) for the inner CV. There, different values for the regularization strength are tested on the respective remaining fold, this time \emph{on the same laboratory} as the training data.\newline 
Due to the relatively high computational cost of training many neural networks, we chose the same value for $\lambda_1$ and $\lambda_2$ in \eqref{eq:minimizer_elastic} for faster training, which we will call $\lambda$. This results in a linear growth in the number of tried-out parameter values instead of a quadratic growth, which saves a considerable amount of computation time. In prior \emph{intra-lab} experiments, this was found as a useful heuristic for TIC-normalized data. The optimal value for $\lambda$ was searched for on the logarithmic grid $G=(10^{-5},10^{-4.5},\dots,10^{-2})$, consisting of 7 values in total. In the prior experiments, this was deemed a sensible range, at the ends of which the performance started to drop again. After the inner CV is finished, we do not choose the value for $\lambda$ with the respective highest validation score, but the next highest value in the grid $G$ (if there is one). We chose this approach, because of the induced overoptimism from validating on the same laboratory as the training set. In other words: For the test sets on the outer CV (which stem from a different lab than the training data), we expect to require a more robust model (i.e. a higher value of $\lambda$) than for the validation sets. This approach mirrors strategies like the \emph{one-standard error rule} (cf. e.g. \citep{hastie2015statistical}), which is another popular heuristic for choosing more robust regularization parameters.
For this cohort, a total of $2\cdot 5\cdot 5 \cdot 7 + 2\cdot 5=360$ neural networks were thus trained, taking roughly 25 GPU-days. While this is a long time, the training of a single model just takes between 1.5 and 2 hours. By using hyperparameter choice heuristics such as the one proposed in Figure \ref{fig:parameter_choice_DRR}, the computation time can be greatly reduced.\newline

\begin{figure}
    \centering
    \includegraphics[width=.45\textwidth]{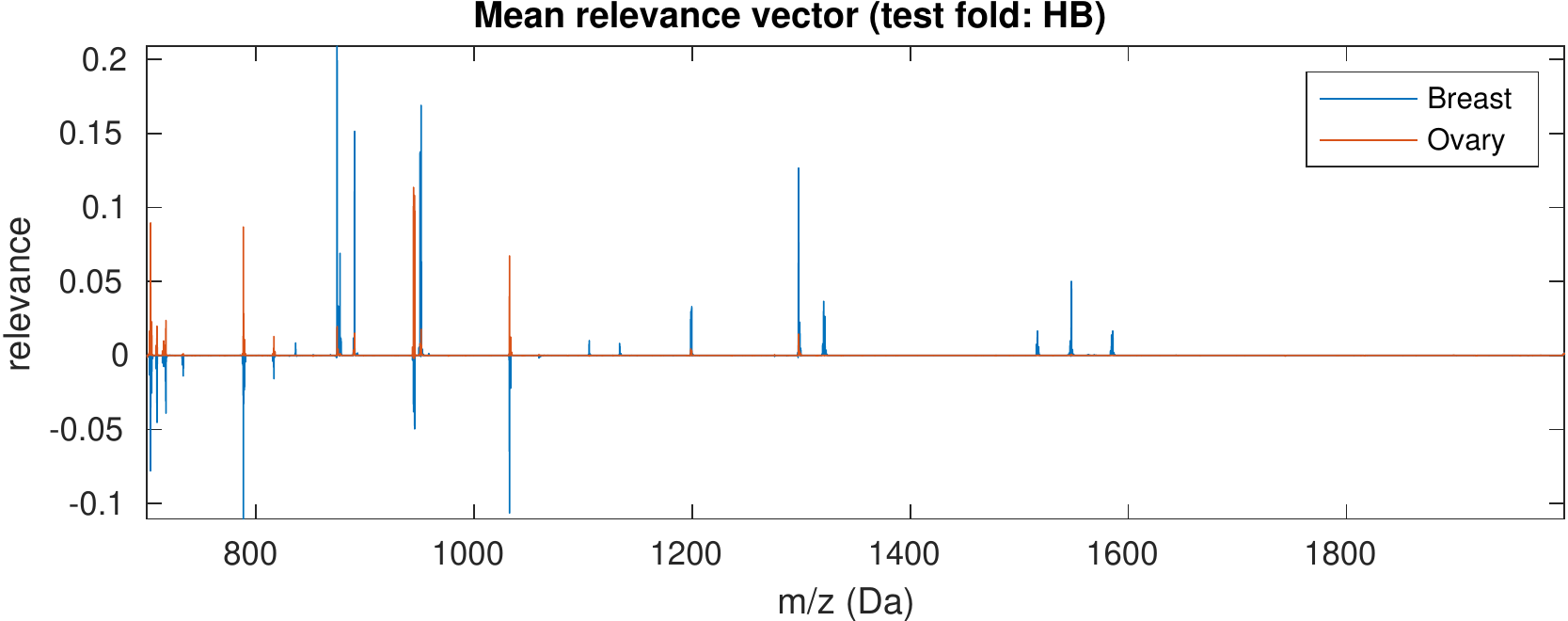}
    \caption{The network trained with deep relevance regularization produces a much sparser mean relevance map on the same test set as the unregularized network in Figure \ref{fig:bad_relevance}. This may mean that actual biomarkers are being learned instead of confounding factors. Interestingly, the trained neural network only seems to collect 'pro-ovary' evidence for the class 'ovary', whereas it collects both evidence and counter-evidence for class 'breast'.}
    \label{fig:good_relevance}
\end{figure}

\noindent\textbf{Results}\newline
\noindent The balanced accuracy for this cohort was 77.4\%, or 80.6\% aggregated on patient-level. This stark increase in accuracy suggests that instead of confounding factors, relevant information is being learned, which persists across measurements. We therefore inspect the mean relevance maps (per class) of the same fold as in the case of the unregularized network and visualize them in Figure \ref{fig:good_relevance}. As desired, these are much sparser than those of the unregularized neural network (Figure \ref{fig:bad_relevance}).

\subsection{Comparison to Linear Model}
We employ the same inter-lab cross-validation procedure for hyperparameter tuning to a comparison model. To be precise, we tune a supervised peak-picking method, whose selected peaks are classified with a linear discriminant analysis classifier (LDA), the same baseline method as in \citep{boskamp2017new,isotopenet,leuschner2018supervised}. For the peak-picking, we select the $k$ m/z-values with the highest \emph{area under the receiver operating characteristic curve} (AUROC) on the respective training set. Like in the case of DRR, the number $k$ is chosen based on the performance in the inner CV. Due to the relatively low computation time, we chose a grid of 16 numbers for $k$ between 5 and 200. Analogously to the DRR-experiment, we chose the next-lowest value on this grid (if possible) compared to the value that gave the best performance on the (intra-lab) validation set. \\

\noindent\textbf{Results}\newline
The above setup leads to a spot-wise balanced accuracy of 75.5\% (78.8\% aggregated over patients). Only taking into account a small subset of m/z-values apparently tends to filter out some of the confounding factors, mirroring the DRR-network approach. Still, the performance of the neural network with DRR is  superior to this linear classification model.

\begin{table}\centering
\caption{Balanced accuracies (spot-wise and patient-wise) for the unregularized NN, the deep relevance regularized NN and the LDA with AUROC feature selection.} \label{tab:balAccs}
\vspace{.2cm}\begin{tabular}{@{}llll@{}} 
\toprule
&NN\hspace{.5cm} & DRR-NN\hspace{.1cm} & ROC/LDA\\  
\midrule 
Spot-level & 37.3\% & 77.4\% & 75.5\%\\
Patient-level \hspace{.2cm} & 34.9\% & 80.6\% & 78.8\%\\
\bottomrule
\end{tabular}
\end{table}

\subsection{Inspection of Relevance Maps}
In the following, we will take a closer look at the relevance maps induced by both deep relevance regularized networks and unregularized neural networks. \\

\begin{figure}\centering
\includegraphics[width=.21\textwidth]{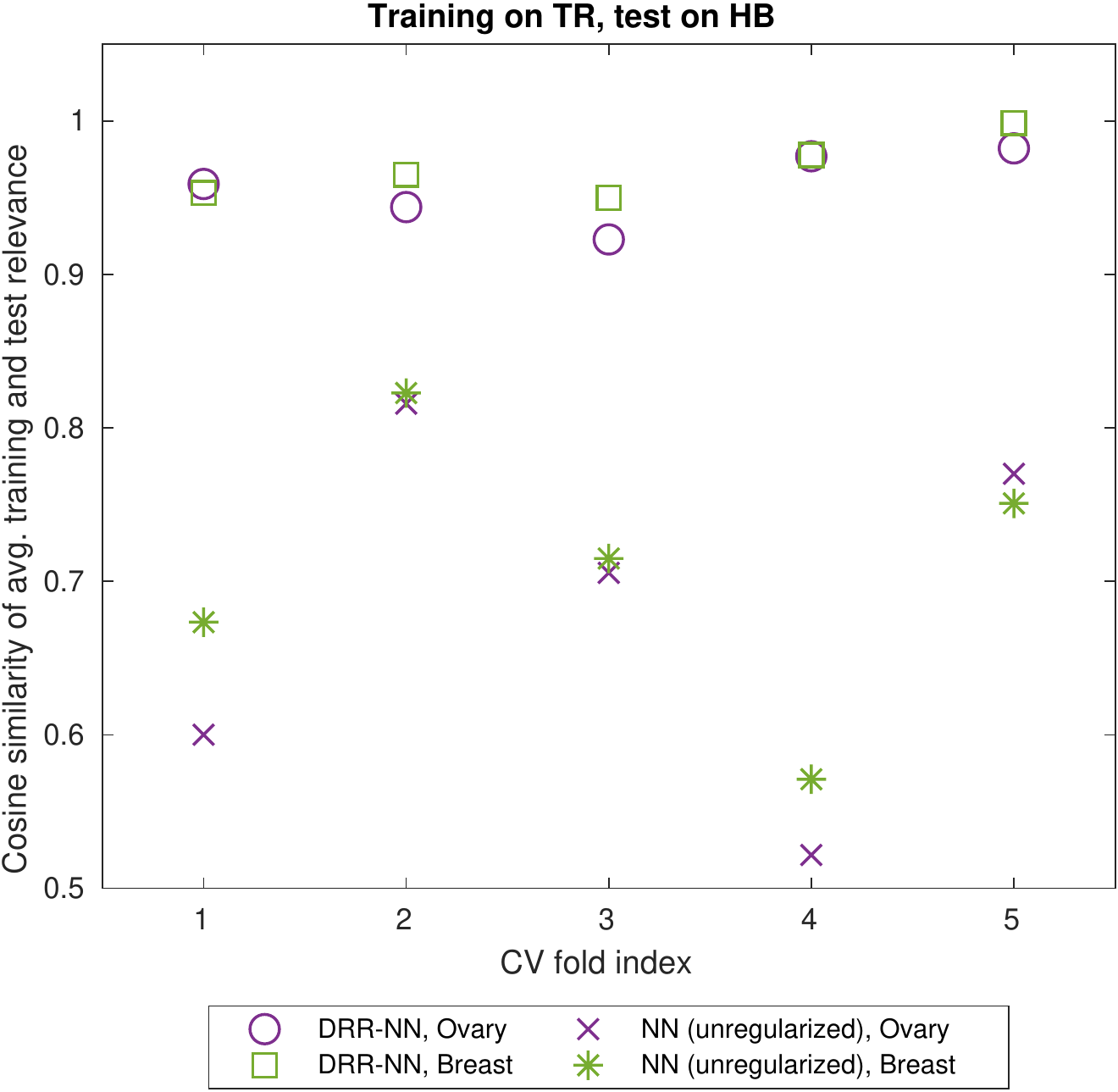}\hspace{.05\textwidth}\includegraphics[width=.21\textwidth]{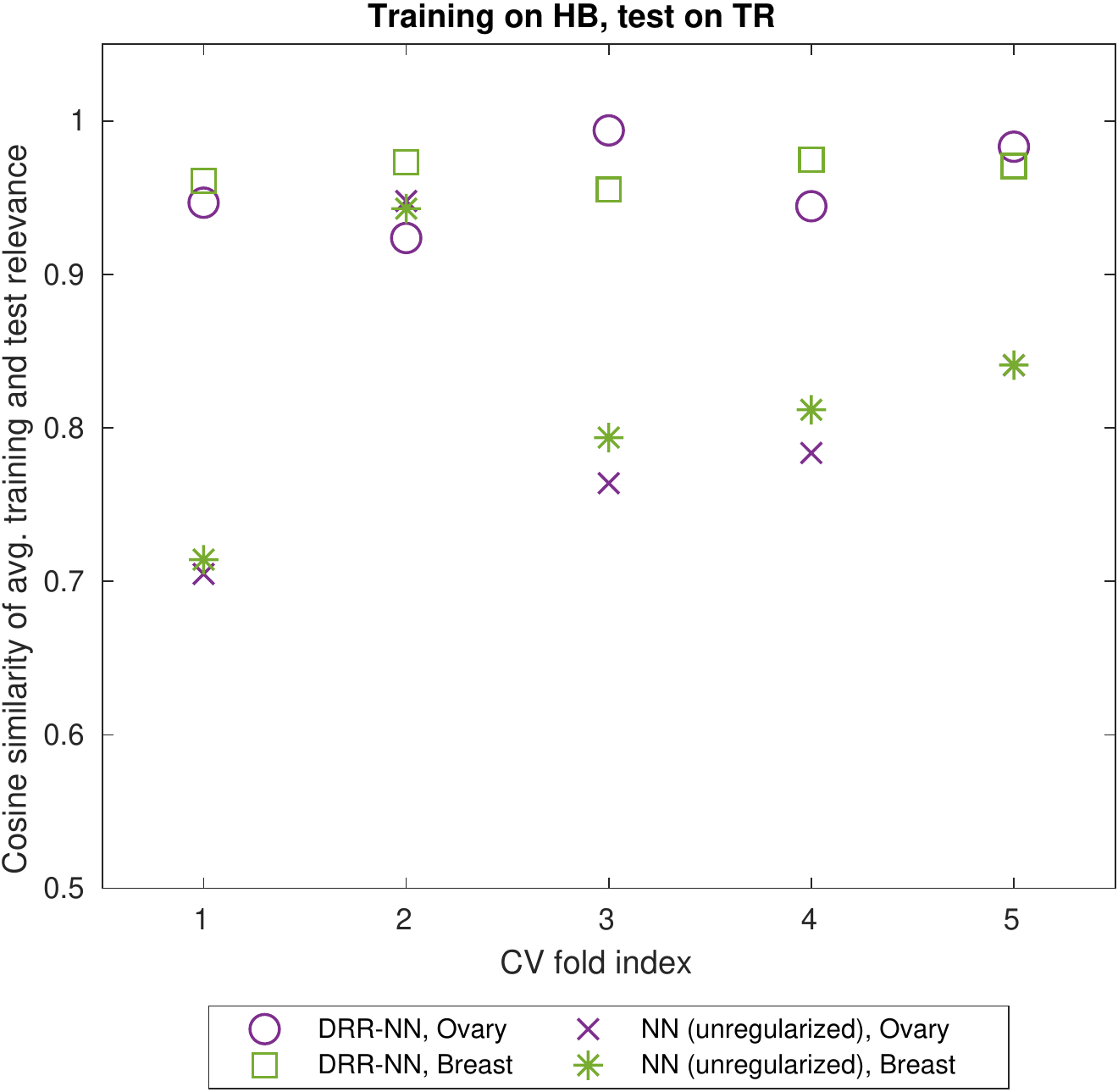}
    \caption{These plots show the cosine similarities between the mean relevance maps of the training sets (from one lab) and the respective test set (from the \emph{other} lab), using the correct labels. Relevance maps of DRR-regularized NNs are considerably more consistent between training and test sets from different labs.}
    \label{fig:relevance_map_comparison}
\end{figure}

\begin{figure}[h]\centering
  \includegraphics[width=.41\textwidth]{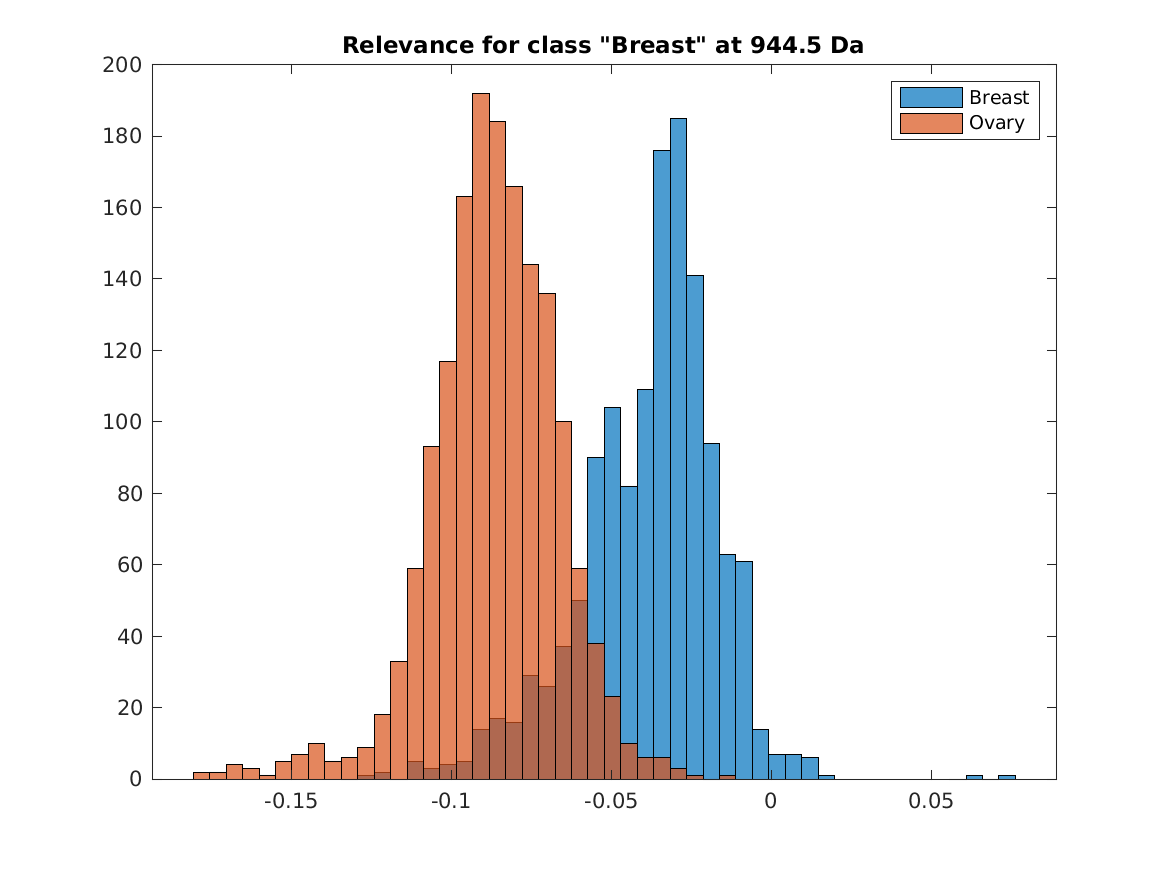}  \includegraphics[width=.41\textwidth]{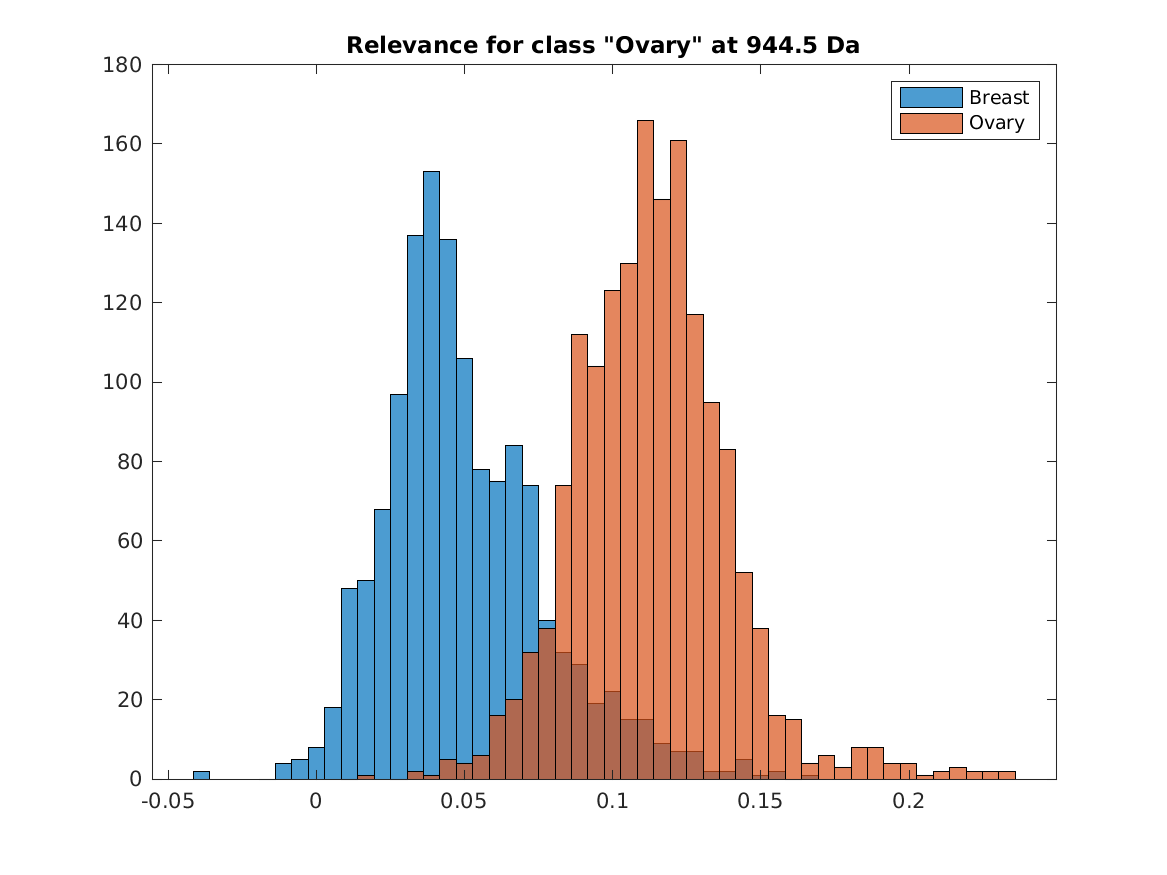}
  \caption{Relevances on the same test fold as in Figure \ref{fig:good_relevance} exemplified for a fixed m/z-value, where the two colors indicate the ground truth. There is a strong correlation between the ground truth class label and the relevance for each class.}\label{fig:relevance_histogram}
\end{figure}

\noindent In Figure \ref{fig:relevance_map_comparison}, we visualize the cosine similarity
\begin{equation}\label{eq:cosine_similarity}
    \cos (u,v) = \frac{u^T v}{\|u\|_2\|v\|_2}
\end{equation}
between the average relevance map for one lab's training folds and the respective test fold of the other lab. The DRR-NNs' relevance maps exhibit a much higher \emph{inter-lab} cosine similarity between training and test sets than their unregularized counterparts. When using an unregularized neural network, it seemingly tends to focus on different portions of the input for unseen data from a different lab. Assuming that relevance maps indeed highlight biologically relevant features, this explains the much better performance when applying DRR.\newline
While we previously examined only average relevance maps, we now look at how consistently these m/z-values are taken into account by the network when using DRR. As exemplified for a high-relevance peak from Figure \ref{fig:good_relevance}, in Figure \ref{fig:relevance_histogram} we observe that the relevances of the chosen peak are quite consistent throughout each class throughout the test set and that they are highly indicative of the ground truth label.

%%%%%%%%%%%%%%%%%%%%%%%%%%%%%%%%%%%%%%%%%%%%%%%%%%%%%%%%%%%%%%%%%%%%%%%%%%%%%%%%%%%%%
%
%     please remove the " % " symbol from \centerline{\includegraphics{fig01-eps-converted-to.pdf}}
%     as it may ignore the figures.
%
%%%%%%%%%%%%%%%%%%%%%%%%%%%%%%%%%%%%%%%%%%%%%%%%%%%%%%%%%%%%%%%%%%%%%%%%%%%%%%%%%%%%%%

\section{Discussion}
In this study, we first identified failures of neural networks for classifying IMS data in the presence of confounding factors induced by inter-lab settings. We did so by employing layerwise relevance propagation, a method for assessing which parts of a mass spectrum contribute most to the classification decision. \newline
Based on this assessment, we proposed a regularization strategy called \emph{deep relevance regularization}, which enforces the classifier to base its decision on few m/z-values. In contrast to classical feature selection approaches, this happens individually for each sample. While a vanilla neural network without deep relevance regularization yields a very bad class-balanced accuracy of 37.3\% on a challenging multi-centric tumor typing task, the thus-regularized version increases this score to 77.4\%, beating a linear comparison method. Both the regularized neural network and the linear method were tuned using an extensive hyperparameter search with a nested inter-lab cross-validation method. The obtained relevance maps of the regularized networks were indeed much sparser than those of their unregularized counterparts, explaining the vastly improved performance.\\

\noindent Still, this work leaves much room for further considerations and possible extensions. Layerwise relevance propagation is just one of many possible attribution methods, as outlined in section \ref{sec:attribution_methods}. While layerwise relevance propagation is both theoretically justified and easy to implement for many models, there may be methods that are more suited to the task of IMS data classification.\newline 
The presented framework of deep relevance regularization also makes the  incorporation of prior knowledge about measurable biomarkers possible: If one expects certain m/z-values (e.g. of specific peptides) to be of importance for the classification task, these m/z-values can simply be excluded (or assigned a lower penalty) than the remaining spectrum. This may be realized using a vector $w$ of weights (e.g. consisting of zeros and ones), resulting in a modified penalty term $\mathcal{R}(\lambda,w \odot \rho_\theta(x,y))$.\newline
Apart from classification, other tasks such as representation learning can be considered. The layerwise relevance propagation of an autoencoder can be equipped with a penalty term, similarly to contractive autoencoders \citep{rifai2011contractive}. By weighting a-priori known m/z-values as proposed above, one would steer the resulting presentation towards a desired task -- e.g. for biologically sensible dimensionality reduction before clustering.

\section*{Acknowledgements}
We thank Jonathan von Schr\"{o}der and Max Westphal for their helpful comments. Tissue samples were kindly provided by Dr. M. Kriegsmann (Institute of Pathology, Heidelberg University Hospital) and Dr. K. Kriegsmann (Department of Hematology, Oncology and Rheumatology, Heidelberg University Hospital) in accordance with the regulations of the ethics committee of Heidelberg University. The authors acknowledge that the method of MS-based differentiation of tissue states are subject to patents held by or exclusively licensed by Bruker Daltonik GmbH, Bremen, Germany.

\section*{Funding}
The authors gratefully acknowledge the financial support from the German Federal Ministry of Education and Research ('KMU-innovativ: Medizintechnik' program, contract number 13GW0081) and the German Science Foundation (DFG) - project number 281474342: 'RTG $\pi^3$: Parameter Identification - Analysis, Algorithms, Applications'.

\bibliographystyle{natbib}
\bibliographystyle{achemnat}
\bibliographystyle{plainnat}
\bibliographystyle{abbrv}
\bibliographystyle{bioinformatics}
\bibliographystyle{plain}
\bibliography{drr_bib}

\begin{thebibliography}{}

\bibitem[Aichler and Walch(2015)Aichler and Walch]{aichler2015maldi}
Aichler, M. and Walch, A. (2015).
\newblock {MALDI} imaging mass spectrometry: current frontiers and perspectives
  in pathology research and practice.
\newblock {\em Laboratory investigation\/}, {\bf 95}(4), 422.

\bibitem[Alberts {\em et~al.}(2018)Alberts, Pottier, Smargiasso, Baiwir,
  Mazzucchelli, Delvenne, Kriegsmann, Kazdal, Warth, De~Pauw, {\em
  et~al.}]{alberts2018maldi}
Alberts, D., Pottier, C., Smargiasso, N., Baiwir, D., Mazzucchelli, G.,
  Delvenne, P., Kriegsmann, M., Kazdal, D., Warth, A., De~Pauw, E., {\em
  et~al.} (2018).
\newblock {MALDI} imaging-guided microproteomic analyses of heterogeneous
  breast tumors—a pilot study.
\newblock {\em PROTEOMICS--Clinical Applications\/}, {\bf 12}(1), 1700062.

\bibitem[Ancona {\em et~al.}(2017)Ancona, Ceolini, {\"O}ztireli, and
  Gross]{ancona2017towards}
Ancona, M., Ceolini, E., {\"O}ztireli, C., and Gross, M. (2017).
\newblock Towards better understanding of gradient-based attribution methods
  for deep neural networks.
\newblock {\em arXiv preprint arXiv:1711.06104\/}.

\bibitem[Bach {\em et~al.}(2015)Bach, Binder, Montavon, Klauschen, M{\"u}ller,
  and Samek]{bach2015pixel}
Bach, S., Binder, A., Montavon, G., Klauschen, F., M{\"u}ller, K.-R., and
  Samek, W. (2015).
\newblock On pixel-wise explanations for non-linear classifier decisions by
  layer-wise relevance propagation.
\newblock {\em PloS one\/}, {\bf 10}(7), e0130140.

\bibitem[Balu {\em et~al.}(2017)Balu, Nguyen, Kokate, Hegde, and
  Sarkar]{balu2017forward}
Balu, A., Nguyen, T.~V., Kokate, A., Hegde, C., and Sarkar, S. (2017).
\newblock A forward-backward approach for visualizing information flow in deep
  networks.
\newblock {\em arXiv preprint arXiv:1711.06221\/}.

\bibitem[Behrmann {\em et~al.}(2018)Behrmann, Etmann, Boskamp, Casadonte,
  Kriegsmann, and Maaβ]{isotopenet}
Behrmann, J., Etmann, C., Boskamp, T., Casadonte, R., Kriegsmann, J., and
  Maaβ, P. (2018).
\newblock {Deep learning for tumor classification in imaging mass
  spectrometry}.
\newblock {\em Bioinformatics\/}, {\bf 34}(7), 1215--1223.

\bibitem[Bojarski {\em et~al.}(2016)Bojarski, Choromanska, Choromanski, Firner,
  Jackel, Muller, and Zieba]{bojarski2016visualbackprop}
Bojarski, M., Choromanska, A., Choromanski, K., Firner, B., Jackel, L., Muller,
  U., and Zieba, K. (2016).
\newblock Visualbackprop: efficient visualization of cnns.
\newblock {\em arXiv preprint arXiv:1611.05418\/}.

\bibitem[Boskamp {\em et~al.}(2017)Boskamp, Lachmund, Oetjen, Hernandez, Trede,
  Maass, Casadonte, Kriegsmann, Warth, Dienemann, {\em et~al.}]{boskamp2017new}
Boskamp, T., Lachmund, D., Oetjen, J., Hernandez, Y.~C., Trede, D., Maass, P.,
  Casadonte, R., Kriegsmann, J., Warth, A., Dienemann, H., {\em et~al.} (2017).
\newblock A new classification method for {MALDI} imaging mass spectrometry
  data acquired on formalin-fixed paraffin-embedded tissue samples.
\newblock {\em Biochimica et Biophysica Acta (BBA)-Proteins and Proteomics\/},
  {\bf 1865}(7), 916--926.

\bibitem[Buck {\em et~al.}(2018)Buck, Heijs, Beine, Schepers, Cassese, Heeren,
  McDonnell, Henkel, Walch, and Balluff]{buck2018round}
Buck, A., Heijs, B., Beine, B., Schepers, J., Cassese, A., Heeren, R.~M.,
  McDonnell, L.~A., Henkel, C., Walch, A., and Balluff, B. (2018).
\newblock Round robin study of formalin-fixed paraffin-embedded tissues in mass
  spectrometry imaging.
\newblock {\em Analytical and bioanalytical chemistry\/}, {\bf 410}(23),
  5969--5980.

\bibitem[Caprioli {\em et~al.}(1997)Caprioli, Farmer, and Gile]{maldi}
Caprioli, R.~M., Farmer, T.~B., and Gile, J. (1997).
\newblock Molecular imaging of biological samples: localization of peptides and
  proteins using {MALDI-TOF MS}.
\newblock {\em Analytical chemistry\/}, {\bf 69}(23), 4751--4760.

\bibitem[Cole(2011)Cole]{cole2011electrospray}
Cole, R.~B. (2011).
\newblock {\em Electrospray and MALDI mass spectrometry: fundamentals,
  instrumentation, practicalities, and biological applications\/}.
\newblock John Wiley \& Sons.

\bibitem[Cordero~Hernandez {\em et~al.}(2019)Cordero~Hernandez, Boskamp,
  Casadonte, Hauberg-Lotte, Oetjen, Lachmund, Peter, Trede, Kriegsmann,
  Kriegsmann, {\em et~al.}]{cordero2019targeted}
Cordero~Hernandez, Y., Boskamp, T., Casadonte, R., Hauberg-Lotte, L., Oetjen,
  J., Lachmund, D., Peter, A., Trede, D., Kriegsmann, K., Kriegsmann, M., {\em
  et~al.} (2019).
\newblock Targeted feature extraction in {MALDI} mass spectrometry imaging to
  discriminate proteomic profiles of breast and ovarian cancer.
\newblock {\em PROTEOMICS--Clinical Applications\/}, {\bf 13}(1), 1700168.

\bibitem[Deininger {\em et~al.}(2011)Deininger, Cornett, Paape, Becker, Pineau,
  Rauser, Walch, and Wolski]{deininger2011normalization}
Deininger, S.-O., Cornett, D.~S., Paape, R., Becker, M., Pineau, C., Rauser,
  S., Walch, A., and Wolski, E. (2011).
\newblock Normalization in {MALDI}-tof imaging datasets of proteins: practical
  considerations.
\newblock {\em Analytical and bioanalytical chemistry\/}, {\bf 401}(1),
  167--181.

\bibitem[Friedman {\em et~al.}(2001)Friedman, Hastie, and
  Tibshirani]{friedman2001elements}
Friedman, J., Hastie, T., and Tibshirani, R. (2001).
\newblock {\em The elements of statistical learning\/}, volume~1.
\newblock Springer series in statistics New York.

\bibitem[Galli {\em et~al.}(2016)Galli, Zoppis, Smith, Magni, and
  Mauri]{galli2016machine}
Galli, M., Zoppis, I., Smith, A., Magni, F., and Mauri, G. (2016).
\newblock Machine learning approaches in {MALDI}-msi: clinical applications.
\newblock {\em Expert review of proteomics\/}, {\bf 13}(7), 685--696.

\bibitem[Goodfellow {\em et~al.}(2016)Goodfellow, Bengio, and
  Courville]{Goodfellow-et-al-2016}
Goodfellow, I., Bengio, Y., and Courville, A. (2016).
\newblock {\em Deep Learning\/}.
\newblock MIT Press.
\newblock \url{http://www.deeplearningbook.org}.

\bibitem[Hastie {\em et~al.}(2015)Hastie, Tibshirani, and
  Wainwright]{hastie2015statistical}
Hastie, T., Tibshirani, R., and Wainwright, M. (2015).
\newblock {\em Statistical learning with sparsity: the lasso and
  generalizations\/}.
\newblock Chapman and Hall/CRC.

\bibitem[He {\em et~al.}(2015)He, Zhang, Ren, and Sun]{He_2015_ICCV}
He, K., Zhang, X., Ren, S., and Sun, J. (2015).
\newblock Delving deep into rectifiers: Surpassing human-level performance on
  imagenet classification.
\newblock In {\em The IEEE International Conference on Computer Vision
  (ICCV)\/}.

\bibitem[He {\em et~al.}(2016)He, Zhang, Ren, and Sun]{he2016deep}
He, K., Zhang, X., Ren, S., and Sun, J. (2016).
\newblock Deep residual learning for image recognition.
\newblock In {\em Proceedings of the IEEE conference on computer vision and
  pattern recognition\/}, pages 770--778.

\bibitem[Kindermans {\em et~al.}(2016)Kindermans, Sch{\"u}tt, M{\"u}ller, and
  D{\"a}hne]{kindermans2016investigating}
Kindermans, P.-J., Sch{\"u}tt, K., M{\"u}ller, K.-R., and D{\"a}hne, S. (2016).
\newblock Investigating the influence of noise and distractors on the
  interpretation of neural networks.
\newblock {\em arXiv preprint arXiv:1611.07270\/}.

\bibitem[Kindermans {\em et~al.}(2017)Kindermans, Sch{\"u}tt, Alber,
  M{\"u}ller, and D{\"a}hne]{kindermans2017patternnet}
Kindermans, P.-J., Sch{\"u}tt, K.~T., Alber, M., M{\"u}ller, K.-R., and
  D{\"a}hne, S. (2017).
\newblock Patternnet and {PatternLRP} -- improving the interpretability of
  neural networks.
\newblock {\em arXiv preprint arXiv:1705.05598\/}.

\bibitem[Kingma and Ba(2014)Kingma and Ba]{adam}
Kingma, D. and Ba, J. (2014).
\newblock Adam: A method for stochastic optimization.
\newblock {\em International Conference on Learning Representations\/}.

\bibitem[Kriegsmann {\em et~al.}(2015)Kriegsmann, Kriegsmann, and
  Casadonte]{kriegsmann2015maldi}
Kriegsmann, J., Kriegsmann, M., and Casadonte, R. (2015).
\newblock {MALDI} tof imaging mass spectrometry in clinical pathology: a
  valuable tool for cancer diagnostics.
\newblock {\em International journal of oncology\/}, {\bf 46}(3), 893--906.

\bibitem[LeCun {\em et~al.}(1989)LeCun, Boser, Denker, Henderson, Howard,
  Hubbard, and Jackel]{lecun1989backpropagation}
LeCun, Y., Boser, B., Denker, J.~S., Henderson, D., Howard, R.~E., Hubbard, W.,
  and Jackel, L.~D. (1989).
\newblock Backpropagation applied to handwritten zip code recognition.
\newblock {\em Neural computation\/}, {\bf 1}(4), 541--551.

\bibitem[LeCun {\em et~al.}(2015)LeCun, Bengio, and Hinton]{lecun2015deep}
LeCun, Y., Bengio, Y., and Hinton, G. (2015).
\newblock Deep learning.
\newblock {\em nature\/}, {\bf 521}(7553), 436.

\bibitem[Lee and Seung(1999)Lee and Seung]{lee1999learning}
Lee, D.~D. and Seung, H.~S. (1999).
\newblock Learning the parts of objects by non-negative matrix factorization.
\newblock {\em Nature\/}, {\bf 401}(6755), 788.

\bibitem[Leuschner {\em et~al.}(2018)Leuschner, Schmidt, Fernsel, Lachmund,
  Boskamp, and Maass]{leuschner2018supervised}
Leuschner, J., Schmidt, M., Fernsel, P., Lachmund, D., Boskamp, T., and Maass,
  P. (2018).
\newblock Supervised non-negative matrix factorization methods for {MALDI}
  imaging applications.
\newblock {\em Bioinformatics\/}, {\bf 35}(11), 1940--1947.

\bibitem[Montavon {\em et~al.}(2017a)Montavon, Lapuschkin, Binder, Samek, and
  M{\"u}ller]{montavon2017explaining}
Montavon, G., Lapuschkin, S., Binder, A., Samek, W., and M{\"u}ller, K.-R.
  (2017a).
\newblock Explaining nonlinear classification decisions with deep taylor
  decomposition.
\newblock {\em Pattern Recognition\/}, {\bf 65}, 211--222.

\bibitem[Montavon {\em et~al.}(2017b)Montavon, Samek, and
  M{\"u}ller]{montavon2017methods}
Montavon, G., Samek, W., and M{\"u}ller, K.-R. (2017b).
\newblock Methods for interpreting and understanding deep neural networks.
\newblock {\em Digital Signal Processing\/}.

\bibitem[Pedregosa {\em et~al.}(2011)Pedregosa, Varoquaux, Gramfort, Michel,
  Thirion, Grisel, Blondel, Prettenhofer, Weiss, Dubourg, Vanderplas, Passos,
  Cournapeau, Brucher, Perrot, and Duchesnay]{scikitlearn}
Pedregosa, F., Varoquaux, G., Gramfort, A., Michel, V., Thirion, B., Grisel,
  O., Blondel, M., Prettenhofer, P., Weiss, R., Dubourg, V., Vanderplas, J.,
  Passos, A., Cournapeau, D., Brucher, M., Perrot, M., and Duchesnay, E.
  (2011).
\newblock Scikit-learn: Machine learning in {P}ython.
\newblock {\em Journal of Machine Learning Research\/}, {\bf 12}, 2825--2830.

\bibitem[Rieger {\em et~al.}(2019)Rieger, Singh, Murdoch, and
  Yu]{rieger2019interpretations}
Rieger, L., Singh, C., Murdoch, W.~J., and Yu, B. (2019).
\newblock Interpretations are useful: penalizing explanations to align neural
  networks with prior knowledge.
\newblock {\em arXiv preprint arXiv:1909.13584\/}.

\bibitem[Rifai {\em et~al.}(2011)Rifai, Vincent, Muller, Glorot, and
  Bengio]{rifai2011contractive}
Rifai, S., Vincent, P., Muller, X., Glorot, X., and Bengio, Y. (2011).
\newblock Contractive auto-encoders: Explicit invariance during feature
  extraction.
\newblock In {\em Proceedings of the 28th international conference on machine
  learning (ICML-11)\/}, pages 833--840.

\bibitem[Rosenblatt(1958)Rosenblatt]{rosenblatt1958perceptron}
Rosenblatt, F. (1958).
\newblock The perceptron: a probabilistic model for information storage and
  organization in the brain.
\newblock {\em Psychological review\/}, {\bf 65}(6), 386.

\bibitem[Ross {\em et~al.}(2017)Ross, Hughes, and Doshi-Velez]{ijcai2017-371}
Ross, A.~S., Hughes, M.~C., and Doshi-Velez, F. (2017).
\newblock Right for the right reasons: Training differentiable models by
  constraining their explanations.
\newblock In {\em Proceedings of the Twenty-Sixth International Joint
  Conference on Artificial Intelligence, {IJCAI-17}\/}, pages 2662--2670.

\bibitem[Sch{\"o}lkopf {\em et~al.}(1998)Sch{\"o}lkopf, Smola, and
  M{\"u}ller]{scholkopf1998nonlinear}
Sch{\"o}lkopf, B., Smola, A., and M{\"u}ller, K.-R. (1998).
\newblock Nonlinear component analysis as a kernel eigenvalue problem.
\newblock {\em Neural computation\/}, {\bf 10}(5), 1299--1319.

\bibitem[Selvaraju {\em et~al.}(2016)Selvaraju, Das, Vedantam, Cogswell,
  Parikh, and Batra]{selvaraju2016grad}
Selvaraju, R.~R., Das, A., Vedantam, R., Cogswell, M., Parikh, D., and Batra,
  D. (2016).
\newblock Grad-{CAM}: Why did you say that? visual explanations from deep
  networks via gradient-based localization.
\newblock {\em arXiv preprint arXiv:1610.02391\/}.

\bibitem[Senko {\em et~al.}(1995)Senko, Beu, and
  McLaffertycor]{senko1995determination}
Senko, M.~W., Beu, S.~C., and McLaffertycor, F.~W. (1995).
\newblock Determination of monoisotopic masses and ion populations for large
  biomolecules from resolved isotopic distributions.
\newblock {\em Journal of the American Society for Mass Spectrometry\/}, {\bf
  6}(4), 229--233.

\bibitem[Shrikumar {\em et~al.}(2016)Shrikumar, Greenside, Shcherbina, and
  Kundaje]{shrikumar2016not}
Shrikumar, A., Greenside, P., Shcherbina, A., and Kundaje, A. (2016).
\newblock Not just a black box: Learning important features through propagating
  activation differences.
\newblock {\em arXiv preprint arXiv:1605.01713\/}.

\bibitem[Simonyan {\em et~al.}(2013)Simonyan, Vedaldi, and
  Zisserman]{simonyan2013deep}
Simonyan, K., Vedaldi, A., and Zisserman, A. (2013).
\newblock Deep inside convolutional networks: Visualising image classification
  models and saliency maps.
\newblock {\em arXiv preprint arXiv:1312.6034\/}.

\bibitem[Springenberg {\em et~al.}(2014)Springenberg, Dosovitskiy, Brox, and
  Riedmiller]{springenberg2014striving}
Springenberg, J.~T., Dosovitskiy, A., Brox, T., and Riedmiller, M. (2014).
\newblock Striving for simplicity: The all convolutional net.
\newblock {\em arXiv preprint arXiv:1412.6806\/}.

\bibitem[Stoeckli {\em et~al.}(2001)Stoeckli, Chaurand, Hallahan, and
  Caprioli]{stoeckli2001imaging}
Stoeckli, M., Chaurand, P., Hallahan, D.~E., and Caprioli, R.~M. (2001).
\newblock Imaging mass spectrometry: a new technology for the analysis of
  protein expression in mammalian tissues.
\newblock {\em Nature medicine\/}, {\bf 7}(4), 493.

\bibitem[Thomas {\em et~al.}(2016)Thomas, Race, Steven, Gilmore, and
  Bunch]{thomas2016dimensionality}
Thomas, S.~A., Race, A.~M., Steven, R.~T., Gilmore, I.~S., and Bunch, J.
  (2016).
\newblock Dimensionality reduction of mass spectrometry imaging data using
  autoencoders.
\newblock In {\em 2016 IEEE Symposium Series on Computational Intelligence
  (SSCI)\/}, pages 1--7. IEEE.

\bibitem[Tibshirani(1996)Tibshirani]{tibshirani1996regression}
Tibshirani, R. (1996).
\newblock Regression shrinkage and selection via the lasso.
\newblock {\em Journal of the Royal Statistical Society: Series B
  (Methodological)\/}, {\bf 58}(1), 267--288.

\bibitem[Zeiler and Fergus(2014)Zeiler and Fergus]{zeiler2014visualizing}
Zeiler, M.~D. and Fergus, R. (2014).
\newblock Visualizing and understanding convolutional networks.
\newblock In {\em European conference on computer vision\/}, pages 818--833.
  Springer.

\bibitem[Zhang {\em et~al.}(2016)Zhang, Lin, Brandt, Shen, and
  Sclaroff]{zhang2016top}
Zhang, J., Lin, Z., Brandt, J., Shen, X., and Sclaroff, S. (2016).
\newblock Top-down neural attention by excitation backprop.
\newblock In {\em European Conference on Computer Vision\/}, pages 543--559.
  Springer.

\bibitem[Zintgraf {\em et~al.}(2017)Zintgraf, Cohen, Adel, and
  Welling]{zintgraf2017visualizing}
Zintgraf, L.~M., Cohen, T.~S., Adel, T., and Welling, M. (2017).
\newblock Visualizing deep neural network decisions: Prediction difference
  analysis.
\newblock {\em arXiv preprint arXiv:1702.04595\/}.

\bibitem[Zou and Hastie(2005)Zou and Hastie]{zou2005regularization}
Zou, H. and Hastie, T. (2005).
\newblock Regularization and variable selection via the elastic net.
\newblock {\em Journal of the royal statistical society: series B (statistical
  methodology)\/}, {\bf 67}(2), 301--320.

\end{thebibliography}

\end{document}